# Surgical Depth Anything: Depth Estimation for Surgical Scenes using Foundation Models

[1]Ange Lou, [2]Yamin Li, [2]Yike Zhang and [1,2]Jack Noble
[1]Department of Electrical Engineering, Vanderbilt University
[2]Department of Computer Science, Vanderbilt University
{ange.lou, yamin.li, yike.zhang, jack.noble}@vanderbilt.edu

**Abstract**: Monocular depth estimation is crucial for tracking and reconstruction algorithms, particularly in the context of surgical videos. However, the inherent challenges in directly obtaining ground truth depth maps during surgery render supervised learning approaches impractical. While many self-supervised methods based on Structure from Motion (SfM) have shown promising results, they rely heavily on high-quality camera motion and require optimization on a per-patient basis. These limitations can be mitigated by leveraging the current state-of-the-art foundational model for depth estimation, Depth Anything. However, when directly applied to surgical scenes, Depth Anything struggles with issues such as blurring, bleeding, and reflections, resulting in suboptimal performance. This paper presents a fine-tuning of the Depth Anything model specifically for the surgical domain, aiming to deliver more accurate pixel-wise depth maps tailored to the unique requirements and challenges of surgical environments. Our fine-tuning approach significantly improves the model's performance in surgical scenes, reducing errors related to blurring and reflections, and achieving a more reliable and precise depth estimation.

## 1. Introduction

Monocular depth estimation (MDE) has emerged as a crucial research area in medical computer vision, especially within the surgical domain [1][2]. By predicting per-pixel depth values, MDE enables advanced technologies such as surgical robotic navigation and dynamic surgical scene reconstruction [3][4]. Currently, most MDE methods depend on Structure from Motion (SfM), which seeks to learn geometric consistency between adjacent frames for depth prediction. However, these methods require high-quality camera motion to accurately predict camera trajectories and maintain geometric consistency. Furthermore, they often necessitate retraining the network for different cases, presenting additional challenges. Additionally, a large amount of high-quality video data is essential for achieving accurate predictions with SfM-based MDE techniques [5].

Obtaining an accurate ground truth depth map during surgical procedures presents inherent challenges, including issues such as reflection, blurring, and bleeding. Recent advancements in foundation models offer a promising solution for robust depth estimation in surgical scenes [6][7]. These models benefit from extensive training on large datasets, which imparts strong zero-shot capabilities. Among these, the Depth Anything model [6] stands out, demonstrating state-of-the-art performance in real-world scenes. Trained on a vast dataset of 63.5 million images (1.5 million labeled and 62 million unlabeled), the Depth Anything model can predict depth maps from a single image. This capability makes it particularly well-suited for the complex and dynamic environment of surgery, especially in scenarios where the camera is stationary, or a stereo setup is not feasible.

Despite being trained on such a large real-world dataset, the Depth Anything model still faces limitations when applied to surgical data. Specifically, specular reflections from tissues and surgical tools can significantly impact its performance. To address these limitations, fine-tuning the model on downstream tasks is necessary. In this study, we collect publicly available surgical datasets to fine-tune the Depth Anything model and compare its performance before and after fine-tuning.

## 2. Method

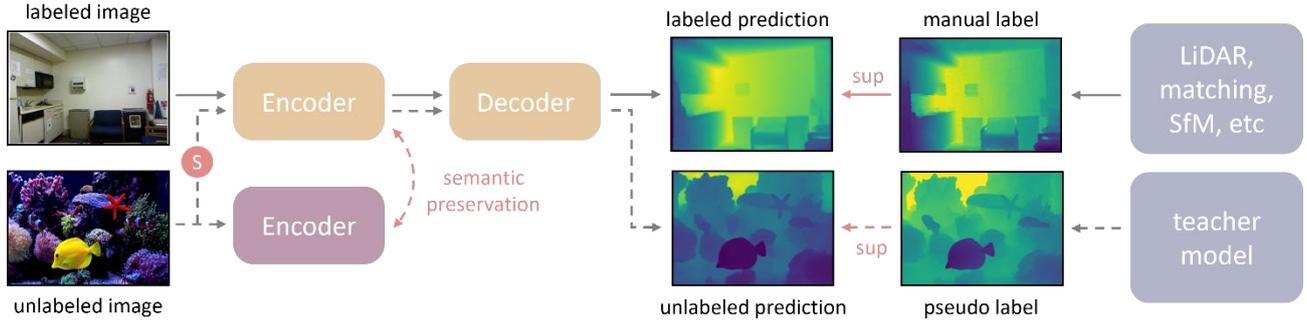

Figure 1. Overall workflow of the Depth Anything model. Solid lines indicate the flow of labeled images, while dotted lines represent the flow of unlabeled images. "S" denotes the application of strong perturbations, including color distortions and CutMix [8]. The online student model is guided to maintain semantic representation through a frozen encoder.

**Depth Anything:** The Depth Anything model employs DINO-V2 [9] as the encoder and DPT [10] as the decoder for the depth prediction task, as illustrated in Figure 1. Initially, a teacher model is trained on a small dataset with ground truth labels and then used to generate pseudo-labels for a large amount of unlabeled data. These pseudo-labels are subsequently used to train a student model, enhancing its ability to predict depth maps accurately. Additionally, a pretrained DINO-V2 encoder is used to preserve the semantic information of the student encoder by calculating cosine similarity in the feature domain.

To improve the performance of the Depth Anything model in the surgical video domain, we fine-tune the model using surgical video data. These surgical datasets include ground truth depth maps, which we feed directly into the Depth Anything model to generate per-pixel depth predictions for each frame. The fine-tuning process is optimized by minimizing the L1 loss between the predicted and ground truth depth maps.

## 3. Experiments and Results

### 3.1 Dataset

To optimize the Depth Anything model for a range of specialized medical tasks, including endoscopic procedures, we utilized the EndoSLAM dataset [11], which comprises three sub-datasets specific to the stomach, small intestine, and colon, and was used to fine-tune the model for endoscopic scenarios. Additionally, the Laparoscopic Image-to-Image Translation (LIIT) dataset [12] was employed to represent different surgical scenarios, including fine-tuning the model for optimal performance in a cochlear implant microscopy task.

The ColonDepth dataset consists of 16,016 RGB images with corresponding ground truth depth maps, all resized to $256 \times 256$ pixels. After fine-tuning the Depth Anything-S model on this dataset, we evaluated its zero-shot performance on the EndoSLAM dataset [2]. The EndoSLAM dataset includes three cases—Stomach, Small Intestine, and Colon—comprising 1,548, 12,558, and 21,887 image-depth pairs, respectively.

For the microscopy task, we fine-tuned the model using the LIIT dataset, which comprises

220,000 images with corresponding ground truth depth maps. We then evaluated the model's performance on our cochlear implant dataset [13].

### 3.2 Measurement Metrics

To quantitatively assess the zero-shot capability of the fine-tuned model, we first apply median scaling, as outlined in Equation (1), to map the relative predictions to the scale of the ground truth.

$$D_{scaled} = D \times \frac{median(\hat{D})}{median(D)} \quad (1),$$

where $D$ and $\hat{D}$ represent the predicted and ground truth images, respectively.

We then compute the absolute relative error and $\delta_1$ score, as defined in Equations (2) and (3).

$$Abs.Rel. = \frac{1}{N}\sum_{i=1}^{N}\frac{|\hat{d}_i - d_i|}{\hat{d}_i} \quad (2),$$

$$\delta_1 = \frac{1}{N}\sum_{i=1}^{N}\left(\max\left(\frac{\hat{d}_i}{d_i}, \frac{d_i}{\hat{d}_i}\right) < 1.25\right) \quad (3),$$

where the $N$ is the total number of pixels, and $d_i$ and $\hat{d}_i$ represent the per-pixel values from the predicted depth $D$ and ground truth depth $\hat{D}$, respectively.

### 3.3 Implementation Details

All experiments were conducted on an RTX A5000 GPU, with a batch size of 16. All inputs were resized to a fixed size of $256 \times 386$. We used the AdamW [14] optimizer with a learning rate of $3e-5$.

### 3.4 Results

We compared the accuracy of relative depth predictions across three versions of the Depth Anything model (small, base, and large) [6], MiDaS [15], and the baseline model from EndoSLAM—Endo-SfMLearner [2]. The standard Depth Anything models and MiDaS are foundational models trained on regular scenes, while Endo-SfMLearner is a self-supervised method specifically trained on monocular surgical videos. Table 1 presents the quantitative results of relative depth estimation on the EndoSLAM dataset. The performance of the standard Depth Anything model, MiDaS, and Endo-SfMLearner is directly sourced from existing studies [16].

Table 1. Quantitative results of relative depth estimation methods on the EndoSLAM dataset. '*' represents the finetuned model and '↑' and '↓' represent the higher is better and lower is better, respectively.

| | Method | Stomach | Small Intestine | Colon |
|---|---|---|---|---|
| **Abs. Rel.↓** | Endo-SfMLearner [2] | 0.438 | 0.474 | 0.551 |
| | MiDaS [15] | 0.309 | 0.351 | 0.354 |
| | Depth Anything-S [6] | 0.354 | 0.585 | 0.616 |
| | Depth Anything-B [6] | 0.284 | 0.567 | 0.607 |
| | Depth Anything-L [6] | **0.239** | 0.574 | 0.626 |
| | Depth Anything-S* | 0.302 | **0.327** | **0.306** |
| **$\delta_1$↑** | Endo-SfMLearner [2] | 0.372 | 0.406 | 0.354 |
| | MiDaS [15] | 0.479 | 0.200 | 0.484 |
| | Depth Anything-S [6] | 0.416 | 0.306 | 0.285 |
| | Depth Anything-B [6] | 0.506 | 0.310 | 0.289 |
| | Depth Anything-L [6] | 0.599 | 0.317 | 0.283 |
| | Depth Anything-S* | **0.637** | **0.622** | **0.622** |

For microscopy surgery, the qualitative results are presented in Figure 2. After fine-tuning on the surgical domain, the Depth Anything model demonstrates an improved ability to capture the intricate structures of the cochlear implant surgical scene. Additionally, the fine-tuned model shows enhanced capability in predicting depth maps that accurately represent fine details within the surgical environment.

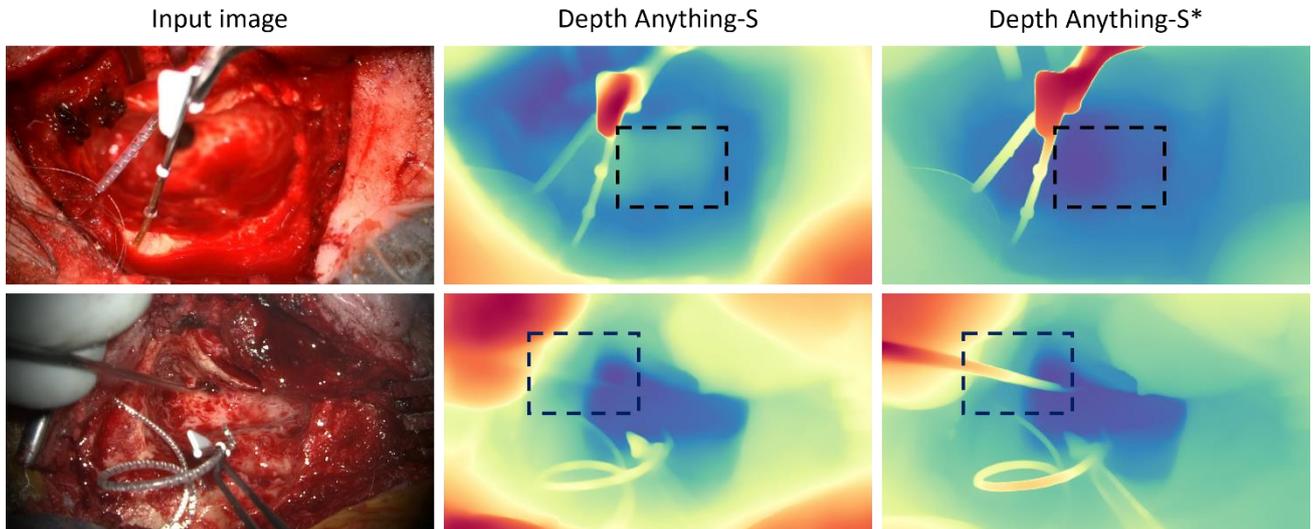

Figure 2. Qualitative results of the cochlear implant surgery. Note: Both the standard and fine-tuned models predict relative depth, so the emphasis is on the structural accuracy of the depth map rather than the exact depth values. Red indicates closer regions, while blue represents farther regions.

## 4. Discussion and Conclusion

Directly applying depth estimation foundation models trained on general scenes poses challenges in accurately predicting the structure of surgical scenes. As shown in Table 1, even after applying median scaling to align predictions with ground truth, these foundation models exhibit significant errors, particularly in the Small Intestine and Colon cases from the EndoSLAM dataset. In our experiments, we fine-tuned the small version of the Depth Anything model. Compared to the model trained on general scenes, the absolute real error was reduced by 15%, 44%, and 50% for the respective cases. Qualitative results, presented in Figure 2, demonstrate that the fine-tuned model provides more accurate depth predictions and better captures fine details in the surgical scenes.

To further evaluate our cochlear implant surgery dataset, we will utilize the predicted depth maps from our fine-tuned Depth Anything model. These depth maps will be fed into a dynamic NeRF model [4] to reconstruct the surgical scene. The fidelity of the reconstructed results will be assessed to reflect the quality of the depth maps.

## Breakthrough work to be presented (this work has not been presented elsewhere)

In this study, we demonstrated the effectiveness of fine-tuning the Depth Anything model for robotic surgery tasks. By leveraging specialized datasets and applying targeted finetuning, we achieved significant improvements in depth prediction accuracy, particularly in challenging surgical environments.